\documentclass[sigconf]{acmart}
\renewcommand\footnotetextcopyrightpermission[1]{} 

\usepackage{booktabs} 
\usepackage{amsmath}
\usepackage{caption}
\usepackage{subcaption}

\usepackage[lined, algoruled, algosection, linesnumbered, resetcount]{algorithm2e}
\usepackage{todonotes}

\DeclareMathOperator{\dis}{d}

\AtBeginDocument{%
  \providecommand\BibTeX{{%
    \normalfont B\kern-0.5em{\scshape i\kern-0.25em b}\kern-0.8em\TeX}}}

\copyrightyear{2021} 
\acmYear{2021} 
\setcopyright{acmlicensed}\acmConference[GECCO '21 Companion]{2021 Genetic and Evolutionary Computation Conference Companion}{July 10--14, 2021}{Lille, France}
\acmBooktitle{2021 Genetic and Evolutionary Computation Conference Companion (GECCO '21 Companion), July 10--14, 2021, Lille, France}
\acmPrice{15.00}
\acmDOI{10.1145/3449726.3463171}
\acmISBN{978-1-4503-8351-6/21/07}

\begin{document}
\title{Behavior-based Neuroevolutionary Training in Reinforcement Learning}
\titlenote{\copyright~Stork et al. 2021. This is the author's version of the work. It is posted here for
your personal use. Not for redistribution. The definitive version was published
in {GECCO'21 Companion: Genetic and
Evolutionary Computation Conference Companion Proceedings}, \url{https://doi.org/10.1145/3449726.3463171}."}

\author{Jörg Stork}
\orcid{1234-5678-9012}
\affiliation{%
  \institution{TH Köln}
  \streetaddress{Steinmüllerallee 1}
  \city{Cologne} 
  \country{Germany}
}
\email{joerg.stork@th-koeln.de}

\author{Martin Zaefferer}
\affiliation{%
  \institution{TH Köln}
  \streetaddress{Steinmüllerallee 1}
  \city{Cologne} 
   \country{Germany}
}
\email{martin.zaefferer@th-koeln.de}

\author{Nils Eisler}
\affiliation{%
  \institution{TH Köln}
  \streetaddress{Steinmüllerallee 1}
  \city{Cologne} 
  \country{Germany}
}
\email{nils.eisler@th-koeln.de}

\author{Patrick Tichelmann}
\affiliation{%
  \institution{TH Köln}
  \streetaddress{Steinmüllerallee 1}
  \city{Cologne} 
  \country{Germany}
}
\email{patrick.tichelmann@th-koeln.de}

\author{Thomas Bartz-Beielstein} 
\affiliation{%
  \institution{TH Köln}
  \streetaddress{Steinmüllerallee 1}
  \city{Cologne} 
  \country{Germany}
}
\email{thomas.bartz-beielstein@th-koeln.de}
 
\author{A. E. Eiben}
\affiliation{%
  \institution{Vrije Universitat Amsterdam}
  \city{Amsterdam} 
  \country{Netherlands}
}
\email{a.e.eiben@vu.nl}

\renewcommand{\shortauthors}{J. Stork et al.}

\begin{abstract}
In addition to their undisputed success in solving classical optimization problems, neuroevolutionary and population-based algorithms have become an alternative to standard reinforcement learning methods. 
However, evolutionary methods often lack the sample efficiency of standard value-based methods that leverage gathered state and value experience. 
If reinforcement learning for real-world problems with significant resource cost is considered, sample efficiency is essential. 
The enhancement of evolutionary algorithms with experience exploiting methods is thus desired and promises valuable insights.
This work presents a hybrid algorithm that combines topology-changing neuroevolutionary optimization with value-based reinforcement learning. 
We illustrate how the behavior of policies can be used to create distance and loss functions, which benefit from stored experiences and calculated state values. 
They allow us to model behavior and perform a directed search in the behavior space by gradient-free evolutionary algorithms and surrogate-based optimization. 
For this purpose, we consolidate different methods to generate and optimize agent policies, creating a diverse population.
We exemplify the performance of our algorithm on standard benchmarks and a purpose-built real-world problem. 
Our results indicate that combining methods can enhance the sample efficiency and learning speed for evolutionary approaches.
\end{abstract}

%
%



\maketitle
\section{Introduction}

In the last years, neuroevolution (NE) and population-based methods have shown to be a valuable and scalable alternative to classic approaches in reinforcement learning (RL), as they have shown promising performance on several problems \cite{salimans2017evolution,jaderberg2017population,jung2020population,khadka2018evolution}. 
In particular, they are computationally efficient if the problem itself is fast to evaluate and a large number of parallel evaluations are possible. 
Evolutionary algorithms (EAs) applied to RL often rely on an episode-to-episode fitness evaluation, considering an RL episode's final cumulative reward for selection and updating.
Further, they do not take advantage of the details of individual behavioral interactions and do not leverage from the gathered experiences of state information.
Thus, they often require a significant amount of fitness evaluations to evolve well-performing agents. 
In particular, if artificial neural networks (ANNs) with changing topologies are considered, which employ a large search space~\cite{Stor18d}.
The resulting low sample efficiency is a challenge in real-world problems due to their high costs, as each action can have a considerable duration.

This paper's primary goal is to combine a topology-changing neuroevolutionary algorithm with behavior-based search components to improve the sample efficiency.

Standard value-based and actor-critic policy gradient methods exhaust behavioral information for their updates and are remarkably successful in many different domains \cite{sutton2018reinforcement}.
The combination of evolutionary methods with value-based methods has recently become an active area of research.
Recent methods include hybrid approaches, which collect experiences by an evolutionary part, then apply value-based learning, such as policy-gradients, to selected population members~\cite{khadka2018evolution,khadka2019collaborative}.
The use of experience-based mutation operators, e.g., which perform gradient updates, is also promising to improve evolutionary methods~\cite{franke2019neural}.

Another approach for improving the sample efficiency is to replace the actual problem function with a surrogate, as employed in surrogate model-based optimization (SMBO). The implementation of surrogates in neuroevolutionary algorithms for reinforcement learning has been of increased interest \cite{gaier2018data,Stor18d}. The difficulty in modeling agents is the definition of an adequate distance between their policies. In the case of changing topologies, the definition of genotypic differences (i.e., differences of the encoding) is algorithm-dependent and not necessarily helpful \cite{gaier2018data,Stor18c}. 
One method is to utilize behavior-based distances, in this context defined as the actions of an agent taken in pre-defined states. Behavioral embeddings are also used to maintain diversity in a population, which is essential to avoid overfitting to certain behaviors~\cite{parker2020effective}. 

In this paper, we seize these ideas and implement a hybrid framework for \emph{behavior-based neuroevolutionary training} (BNET), which combines the evolution of a population of agents with topology-changing neuroevolution, employed by \emph{cartesian genetic programming} (CGP)~\cite{miller2000cartesian,turner2015introducing}.
Our contributions are as follows: 

\textbf{(1)} We introduce the new algorithm where the population's agents are optimized in parallel by fitness-based search, behavior-based search, and surrogate-based search.
As a critical challenge in RL is exploration, our implementation is focused on maintaining a diverse set of agents, each contributing to the fitness search and sharing experiences. 
Different approaches to leverage from a shared experience set exist~\cite{schmitt2020off}; we employ a selection method to create an archive of experiences from high-performing yet diverse policies. 

\textbf{(2)} We define a set of new advantage-weighted behavior loss functions to conduct the behavior-based search, leveraging on principles from standard actor-critic policy algorithms and behavior-based distances~\cite{sutton2018reinforcement,Stor18c}. 
In combination, they are part of a gradient-free learning process for the agent policies. For the surrogate-based search, we employ the \emph{surrogate model-based optimization for neuroevolution} (SMB-NE) algorithm, first introduced in \cite{Stor18d}.

\textbf{(3)} We define a method for robust candidate selection to increase learning stability. Our fitness-based search focuses on keeping a stable elitist solution, referred to as champion, which is of central importance considering real-world environments.
As these environments frequently provide stochastic rewards, e.g., caused by random starting conditions, we employ a robust selection method, which compares mean estimates of their actual performance and reduces the probability of choosing an inferior candidate.

\textbf{(4)} We implement a prototype of the framework and test it on common synthetic benchmark problems against standard value-based RL methods to prove our concept. We also analyze the performance of each search method in BNET. 
Finally, we introduce a new adaptable real-world problem featuring a robot-controlled maze environment as a benchmark for RL algorithms.


\section{Methods}\label{sec:methods}
\subsection{Reinforcement Learning and Value-based Methods}\label{ssec:reinf}
In RL, we train agents' policies $\pi_{n}, n=1,...,N$ to solve instances of an environment: 
for each environment step $t$ the agent observes an environment state $s_t$ and the policy decides, which action $a_t$ is conducted. 
The agent receives a reward $r(s_t,a_t,s_{t+1})$ for each observed state during its learning episode.
This leads to a \textit{trajectory} with $T$ steps. 
The cumulative reward $R_t =\sum_{k=0}^T \gamma^k r_{t+k+1}$ at the end of an episode, discounted by factor $\gamma$,
 is typically utilized as a target function of the policy optimization process.
We refer to $R_t$ of a full episode as \textit{fitness} of a policy.

Advantage-based actor-critic methods~\cite{sutton2018reinforcement} have a policy actor and estimate the \textit{advantage} of an action $a_t$ in state $s_t$ by a critic.
The critic $V_\varphi (s_t)$, with $\varphi$ being its parameters, approximates the estimated \textit{value} of a state $s_t$, given by the value function $V(s)=\mathbb{E}_{\pi} \{R_t | s_t=s\}$. 
For a full-episode learner, i.e., considering the complete per-state reward information $R_t$ at the end of an RL episode, a typical approach is to compute the advantage by a Monte-Carlo (MC) estimate $A_{\varphi}(s_t,a_t)= R_t(s_t,a_t) - V_{\varphi}(s_t)$,
where $V_{\varphi}(s_t)$ is approximated by the critic. 
Further, the actor is updated with help of a gradient function $\nabla_{\theta} J(\theta)=\sum_t \nabla_{\theta} log \pi_{\theta}(s_t,a_t)A_{\varphi}(s_t,a_t)$
Here $\theta$ are the ANN parameters, and $\pi_{\theta}$ is the policy associated with these parameters. 
We refer to the \textit{experience} of an agent, considering the $(s_t,a_t,r_{t+1})$ triplets an agent observes during (multiple) interaction episodes in a RL environment.

\subsection{Neuroevolution by Cartesian Genetic Programming}
CGP~\cite{miller2000cartesian} is a genetic programming method relying on grid-based encodings to realize graph representations. 
If applied to ANNs~\cite{turner2015introducing}, it allows to create topologies with different transfer functions and free node-to-node connections, i.e., the generated topologies do not follow the typical layered structure. 
CGP, in its basic version, does not allow the direct use of back-propagation or gradients for ANN optimization, mainly due to the topology-changing neuroevolution process. 
Certain CGP implementations allow the utilization of gradient information~\cite{izzo2017differentiable}. 
However, these learn topologies and weights sequentially, not simultaneously.
The genotype-behavior mapping is complicated as distances on the genotype-level are barely 
related to their distance in behavior space~\cite{Stor18c}. 
We optimize the CGP-ANNs  with a gradient-free ($\mu + \lambda$)-EA with rank-based fitness selection~\cite{eiben2003introduction} for the optimization of neuroevolution candidates, utilizing behavior-based loss functions, as described in Section \ref{sec:behoptim}.

\subsection{Optimization in the Behavior Space}
\label{sec:behoptim}

Each policy computes the probability $p(a_t|s_t)$ of taking an action $a_t$ for an input state $s_t$. 
Formally, we denote \emph{behavior} as the set of probabilities corresponding to $K$ states $\mathbf{S}=\{s_{1},...,s_{K}\}$. 
Further, $\mathcal{B}$ is denoted as the \textit{behavior space}, i.e., the set of all possible behaviors,
with $\pi_{n} \in \mathcal{B}$ and $n=1,...,N$.
In the behavior space, two agents can be directly compared on the state set $\mathbf{S}$ by calculating their mean \textit{behavior distance}~\cite{Stor18c,Stor18d}, denoted by
\begin{equation}
\dis_b(\pi,\pi',\mathbf{S})= \frac{1}{T} \sum_{t=1}^T | \pi(\mathbf{s}_t) - \pi'(\mathbf{s}_t) |
\end{equation}

For the case of experience-based policy optimization, we defined the \textit{advantage-weighted behavior distance}:
\begin{equation}
\label{eq:awbd}
\dis_{wb}(\pi,\pi',\mathbf{S},\mathbf{A})= \bigg( \frac {\frac{1}{T^*} \sum_{t=1}^{T^*} | \pi(s_t) - \pi'(s_t) | \times A_{\varphi}(s_t, a_t)}{ \frac{1}{T^*} \sum_{t=1}^{T^*}|A_{\varphi}(s_t, a_t)|} \bigg )
\end{equation}
Here, $\mathbf{S}$ are sampled states from environment interactions with a trajectory of $T^*$ time-steps, while $\mathbf{A}$ are the respective actions taken during this trajectory. 
Further, $A_{\varphi}$ is the precomputed approximated advantage for each of the stored actions, see sec.~\ref{ssec:reinf}.
As a second method, we utilize the \textit{advantage-weighted cross-entropy}. The cross-entropy is defined by:
\begin{equation}
H(\pi,\pi',s_t)= - \sum_{i=1}^I \big[ \pi_i(s_t) \ln \big( \pi'_i(s_t) \big) \big]
\end{equation}
$\pi_i(s_t)$ is the i-th element of the probability output of the policy $\pi$ given the input state $s_t$. 
The cross-entropy distance ($d_c$) is further weighted by the advantage estimation $A^+_{\varphi}(a_t, s_t)$ and computed over a set of stored states $\mathbf{S}$ and actions $\mathbf{A}$:
\begin{equation}
\label{eq:awce}
d_{c}(\pi,\pi',\mathbf{S},\mathbf{A})=\frac{1}{T^*} \sum_{t=1}^{T^*} \big[ H(\pi,\pi',s_t) \times A^+_{\varphi}(s_t, a_t) \big]
\end{equation}
Instead of the complete estimated advantage, we only consider the set of positive advantages: 
\begin{equation}
 A^+_{\varphi}(s_t, a_t)= 
\begin{cases}
 A_{\varphi}(a_t, s_t),& \text{if } A_{\varphi}(s_t, a_t) \geq 0 \\
 0,    & \text{otherwise}
\end{cases}
 \end{equation}
 
The positive advantages drive the search towards estimated beneficial behavior, while negative advantages will not affect the weighted cross-entropy distance $d_c$. 
We define our loss function for NE-EA during the behavior-based optimization as the sum of distances ($d_{wb}$ or $d_c$) between the target policy $\pi$ and several stored reference behaviors $\pi_m$ with $m=1,...,M$.
Furthermore, their sampled trajectories with states $\mathbf{S}_m$ and actions $\mathbf{A}_m$ are:
\begin{equation}
\label{eq:loss}
L(\pi)= \sum_{m=1}^M \big[ \dis(\pi,\pi_m,\mathbf{S}_m,\mathbf{A}_m) \big].
\end{equation}
For the loss functions, the probability distribution of the stored behaviors is priorly adapted to maximize the performed action's probability.
The optimization in the behavior space $\mathcal{B}$ is similar to imitation learning, i.e., fitting a network to replicate a stored behavior.
However, instead of replicating a single reference behavior, we optimize the target policy to minimize the distance to an advantage-weighted set of multiple reference behaviors from different policies. 
The EA for the optimization is outlined in algorithm \ref{alg:neuroevolution}. 

\begin{algorithm}[hbtp]
 \caption{Neuroevolution EA}
 \label{alg:neuroevolution}
 \textbf{INPUT:} Memory of $M$ stored reference policies $\pi^*_m$, states and actions $\mathbf{S}^*_m,\mathbf{A}^*_m$; \\
 \textit{optional: pre-defined candidates $\pi$}    \\
 \textbf{preset:} \emph{mutation rate}, \emph{NE candidate parameters} \\
 \Begin{
 \textbf{initialize} new polices as candidates \\
 \textbf{evaluate} initial candidates with loss function $L(\pi)$\\
  \textbf{select} $\mu$ parents from initial candidates \\
 \While{\textbf{not} termination-condition}{
 \textbf{mutate} parents to get $\lambda$ offspring  \\
 \textbf{evaluate} offspring with loss function $L(\pi)$  \\
 \textbf{select} $\mu$ next iteration's parents with minimum loss from parents and offspring \\
 \emph{\textbf{optional:} update mutation rate} \\
  }}
  \textbf{OUTPUT:} best found policies 
\end{algorithm}

\subsection{Optimization by Behavior Surrogates}
\label{sec:surr}
The definition of the behavior distance between policies also allows us to create approximation models, so-called surrogates, which predict a policy's fitness during optimization. 
The surrogates allow searching in the behavior space without any additional environment interactions, which are substituted by the surrogate.
Surrogate model-based optimization (SMBO) is frequently applied for costly processes with high resource-demand for each function evaluation~\cite{Forrester2008a}.
Our SMBO is based on the surrogate model-based optimization for the neuroevolution (SMB-NE) algorithm \cite{Stor18d}, which utilizes a Kriging~\cite{Forrester2008a} regression model.
The model measures the similarity of samples by a kernel, utilizing distance and correlation matrices of observations.
If applied to real-valued samples, the exponential kernel $\text{k}(x,x')=\exp(-\theta ||x-x'||_2)$ is a typical choice. 
The essential kernel parameter $\theta$ influences how fast the correlation decays to zero if the Euclidean distance between two samples $||x-x'||_2$ increases.
In our work, we follow the idea of kernel-based models for combinatorial search spaces~\cite{Moraglio2011,Zaefferer2014b}. 
We replace the Euclidean distance $||x-x'||_2$ by the \textit{behavior distance}, resulting into the kernel:
\begin{equation}
\label{eq:kernbd}
\text{k}(\pi,\pi',\mathbf{S})=\exp(-\theta \dis(\pi(\mathbf{S}),\pi'(\mathbf{S})))
\end{equation}
Here, one challenge is the appropriate definition of the state set $\mathbf{S}$, 
as it has a considerable influence on the distance.
We rely on a selection approach presented in \cite{Stor20a}, 
where both policies' stored states are combined for each pairwise distance calculation. 
If a new target network's fitness without stored states needs to be predicted, the reference policies' stored states are applied as reference input. 
The Kriging model parameters are fitted by maximum likelihood estimation and optimized with DIRECT-L \cite{gablonsky2001locally}, further utilizing the \emph{nugget effect} \cite{van2003kriging}.
Kriging combines relatively accurate mean predictions with the ability to provide uncertainty estimates of each prediction.
The combination of mean and uncertainty are used to compute infill functions, which predict the desirability of a solution. 
A frequently applied infill-criterion is Expected Improvement (EI)~\cite{Mockus1978,Jones1998}, 
which integrates the uncertainty estimate to explore new solutions, which may be farther away from observed solutions.
In \cite{rehbach2020expected}, the benefits and downsides of using EI for different optimization problems are discussed.
For high-dimensional cases, it is recommended to use the predicted mean instead of EI since the increase in dimensionality leads to an inherent increase in uncertainty (see also: \cite{wessing2017true}).
In the case of neuroevolution, the behavior space is high-dimensional.
Thus, we use the predicted mean of the Kriging surrogate as an infill function. 
The R package \texttt{CEGO}~\cite{CEGOv2.2.0,Zaefferer2014b} is used to train the Kriging surrogates, 
while the NE-EA is utilized to optimize the target network.
\section{Behavior-based Neuroevolutionary Training}
\label{sec:bnet}

\begin{figure*}[ht]
\centering
\includegraphics[width=0.65\textwidth]{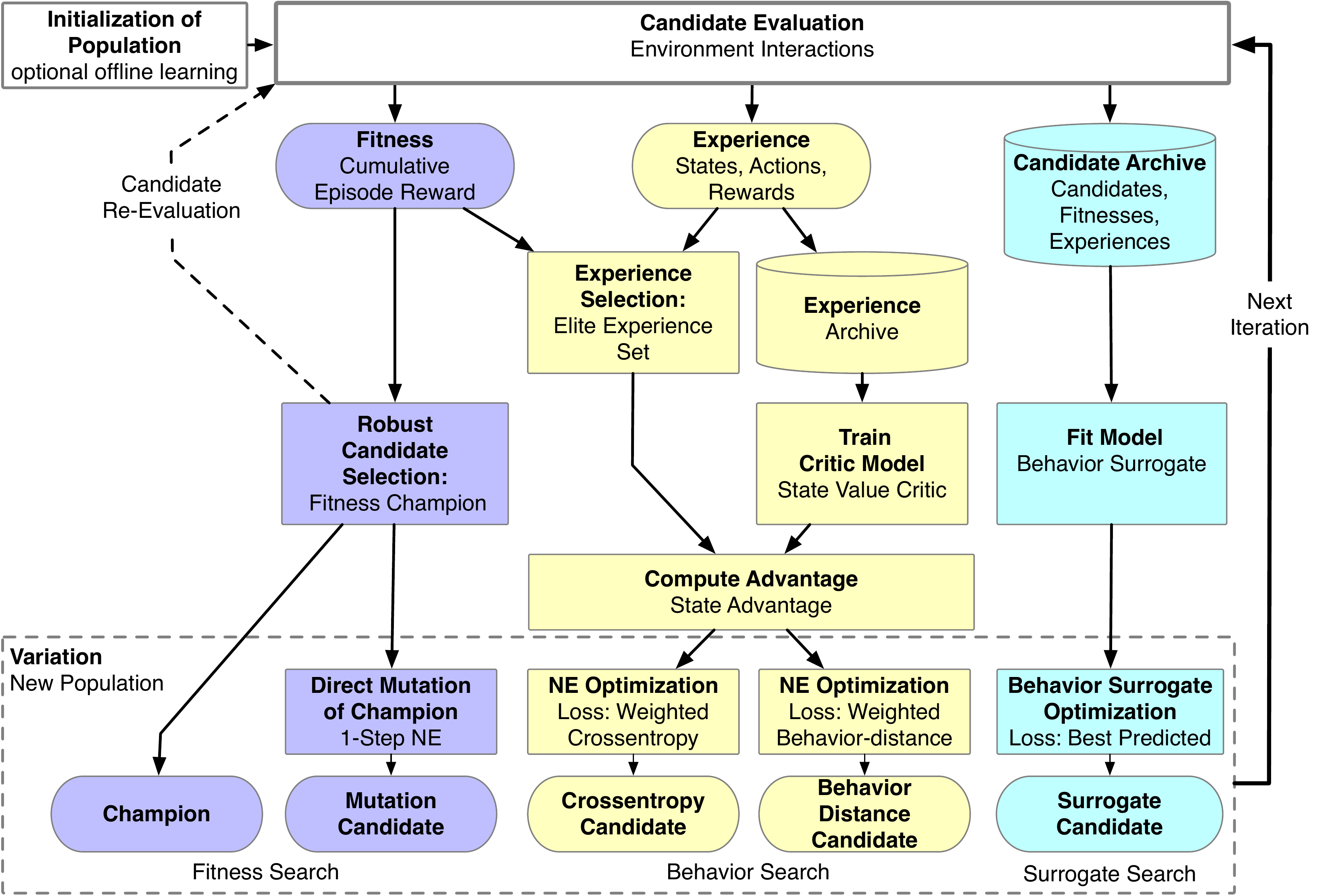}
\caption{BNET algorithm cycle. Quadratic boxes are methods or algorithms, and rounded edges illustrate observed data.
The candidate generation methods are based on extracted data from the environment interactions during the fitness evaluation: The fitness-based search selects a champion with the best overall fitness (violet). The behavior-based search utilizes a selected set of experiences, and advantage-critic model, and the defined behavior-distance and cross-entropy loss-functions to optimize networks and use as new candidates (yellow). Further, a surrogate model is fitted to all candidates' archives and generates a candidate in a surrogate-model-based search (light-blue). }
\label{fig:bnet}
\end{figure*}

In this section, we introduce our new algorithm for population-based neuroevolution for the training of RL policies (BNET). 
The population consists of CGP-ANNs, which are utilized as RL policies $\pi$.
We refer to each network instance as a candidate. 
BNET is similar to a population-based evolutionary algorithm with elitism. 
Nevertheless, new candidates are not merely created employing a variation of previous candidates.
Instead, new candidates are also generated by behavior learning and by exploiting behavior-based surrogates.
The algorithm outlined in Fig. \ref{fig:bnet} is essentially divided into three parts, 
which employ different ways to search for new candidates: 
fitness-based, behavior-based, and surrogate-based. 
In summary, the fitness-based search selects a champion with the best overall fitness by robust selection and creates direct mutations of this champion. 
The behavior-based search utilizes a selected set of experiences, a value-based critic model, and the loss function from Eq. \ref{eq:loss} to generate new candidates. 
Further, a surrogate model is fitted to an archive of all candidates and searched for new candidates.

\subsection{Initialization and Evaluation}
The first population is either initialized by NE or by optimizing a population to fit previously evaluated policies' behavior.
We refer to the second method as \textit{offline} initialization, as it does not require any new (online) environment interactions. 
If we initialize NE policies offline, the behavior search (as explained below) is applied, where a previously-stored experience set is required as a reference.

The candidates are then evaluated in the RL environment with two goals:
First, evaluate the current population's policies to acquire their fitness; Second, to gather new environment experiences from these policies.
The policies can be evaluated either fully deterministic, stochastic, or stochastic with exploration. 
The deterministic policy evaluation chooses the action with the highest probability, whereas the stochastic evaluation samples actions based on the probabilities. 
Deterministic policies have equal behavior if evaluated repeatedly, but the achieved fitness still may differ, e.g., if the environment itself is stochastic.  
Deterministic policies are most suitable for exploitation or testing.
As stochastic policies depend on probabilities, they allow for a certain level of exploration. 
Additional exploration can be enforced by taking random actions or adding a random factor to any policy behavior.

\subsection{Fitness Search and Robust Selection}
The fitness search, visualized as the leftmost, violet path in Fig. \ref{fig:bnet}, focuses on selecting a \textit{fitness champion}, 
i.e., the candidate of the population with the best overall fitness. 
Moreover, a mutation candidate is generated by applying a small direct mutation to the champion's policy network and used as a second new candidate, the \emph{mutated champion}.
However, RL environments, such as control tasks, typically have different starting conditions or even stochastic state transitions, i.e., an action in a specific state may transition to different states in the next time step. 
Thus, the measured fitness will be noisy, and even a deterministic policy produces different results when evaluated repeatedly.

Therefore, we employ a robust fitness selection, including the re-evaluation of candidates and comparing their fitness distributions. 
The robust selection shall ensure that the champion is a reasonable estimate of the best policy discovered so far, even in the presence of noisy fitness. 
In the first iteration, the fitness champion, denoted as $\pi^*$, is selected as the candidate with the highest sampled fitness so far and re-evaluated in the second iteration.
Its fitness is then set to the mean of all evaluations. 
Starting with the second and all future iterations, we use the concept of challengers: 
A challenger is a candidate $\pi$ that has a one-time evaluated fitness better than the mean champion fitness, i.e., if $f(\pi) > \frac{1}{n^*} \sum_1^{n^*} f(\pi^*)$, where $n^*$ is the number of samples of champions performance $f(\pi^*)$.
If a new population contains at least one challenger, this challenger is re-evaluated to get a more robust fitness estimate and
the mean fitness values are utilized as a measure for selection. 
A challenger is accepted as champion, if $\frac{1}{n} \sum_1^{n} f(\pi) > \frac{1}{n^*} \sum_1^{n^*} f(\pi^*)$, where
$n$ is a parameter of the selection, either set to match $n^*$ or a desired number of repeats $r$, by $n= \max(n^*,r)$.
The value of $r$ should be chosen according to the task and estimated noise of the environment.
If a champion was re-evaluated less than $r$, it is also re-evaluated. 

In case of multiple challengers, they are ordered by their fitness and sequentially evaluated against the champion. 
If a new champion is selected during the consecutive comparisons, the next challenger is only considered if it is still superior in fitness. 
The selected champion is then re-evaluated in the next iteration. 
The constant re-evaluation of the champion, if not changed, leads to a stable estimate of the actual fitness value. 


\subsection{Behavior Search}
The behavior-based optimization of the neuroevolutionary policies builds upon principles from standard RL algorithms, such as \emph{policy-gradients} and \emph{actor critic}. 
The challenge in neuroevolution is the absence of any gradient information between NE candidates if a simultaneous topology optimization is performed. 
A neuroevolutionary behavior optimization thus requires metrics that allow the comparison of policies further to establish a search direction in the behavior space. 
We employ the behavior optimization with the \textit{advantage-weighted behavior distance} from Eq. \ref{eq:awbd} or the \textit{advantage-weighted cross-entropy} Eq. \ref{eq:awce} and
implement it in our loss function Eq. \ref{eq:loss}.
This loss application requires a defined set of experiences, an advantage function, and an optimization algorithm. 
As a reference set, we collect and store a fixed-sized set of \textit{elite experiences}.
The set consists of experiences and the performance of single evaluations from different candidates or repeated evaluations of the same candidate. 
In the first iterations, the set grows until the maximum size of the set is reached. 
After the robust selection, the experience is replaced with those of higher fitness episodes in each consecutive iteration. 
However, the number of replacements in each iteration is limited.
The limitation prevents that the candidate set gets dominated by the experiences of single candidates.
A diverse set is thought to help avoid overfitting and increase the chance to learn better behavior. 
A diverse experience set is the first difference to classic imitation learning, where a single policy's behavior is adapted. 
The second difference is given by the advantage weighting of each action in the distances by Eq. \ref{eq:awbd} and \ref{eq:awce}.
The required advantage function is predicted by a critic model, which is learned to the complete set of all discovered experiences.
Also, the advantage-weighted cross-entropy considers only the actions with positive advantage, i.e., the policy is trained to replicate this behavior. 
Both loss metrics are employed in the NE-EA to generate new candidates. 

\subsection{Surrogate Search}
Corresponding to section \ref{sec:surr}, the surrogate search is based on a Kriging model fitted to an archive of tested candidates with their connected mean fitness and experiences.
 The distance kernel in Eq. \ref{eq:kernbd} is applied for modeling the relations between policy behaviors and their fitnesses, where the state set $\mathbf{S}$ of each comparison consists of the stored experience archive of the associated candidate pair. 
The fitted surrogate is then employed to predict the fitness of new candidates and utilized as loss-function $L(\pi)= -\hat{f}(\pi)$ during a NE-EA optimization process. 
The fitness values are adapted to ensure a minimization problem (i.e., negated in the typical reward maximization case).

\section{Experiments}
\label{sec:experiments}
The BNET framework is flexible, as the algorithm modules for generating candidates can be combined in several ways.
For example, it can also be employed as a pure direct neuroevolution approach by refraining from using the behavior-based or surrogate-based search, or in contrast, as a pure behavior search algorithm. 
Thus, our experiments are two-fold: first, we tested different versions of the algorithm against a set of open-AI baselines algorithms on the problems CartPole-V0 and MountainCar-V0. For this experiment, the focus was to estimate overall performance, how beneficial the different proposed modules are, and how they affect the search quality.
In a second experiment, we tested our algorithm against the same baselines on a newly designed real-world problem.

As comparison baselines, we employ \emph{Advantage Actor Critic} (A2C) \cite{mnih2016asynchronous} and \emph{Proximal Policy Optimization} (PPO) \cite{schulman2017proximal} from the \emph{stable baselines} package~\cite{stable-baselines}. 
Both do not require full episodes to learn (i.e., the algorithm is trained after x time steps, not necessarily full episodes), which might give it performance advantages over BNET. 
Our performance measure, particularly concerning the real-world environment, is the number of required time steps until a (stable) solution is found. 
For all experiments, we implement a prototype version of the BNET framework using R 3.6.3, an R-interface to the CGP-ANN Library by Turner \cite{Stor18d} and \emph{reticulate 1.14}, \emph{Keras 2.3}, OpenAI \emph{gym 0.18.0}, and \emph{tensorflow 1.15.4}~\cite{brockman2016openai,tensorflow2015-whitepaper,chollet2015keras}.
All simulated experiments were conducted on an HPC-Cluster. 
More than 50,000h of total computation time was spent during development and experimentation.
The BNET prototype was not systematically tuned for optimal parameter settings due to the high computational effort. The used parameter setup is based on preliminary tests and CGP-ANN or SMB-NE related publications~\cite{Stor18d,Stor20a}. 
The baseline algorithms parameter were also improved (from the stable baseline default settings) based on preliminary results for each problem at hand, e.g., the reward discount parameter gamma and the learning rate.
One aspect of the BNET setup was kept equal for all tested problems: The CGP-ANNs have a maximum of 200 active nodes with arity ten and a function set including tanh, sigmoid, gaussian, softmax, step, and rectified linear units. 
The direct, behavior, and surrogate search's mutation rates were 1\%,5\% and 5\%, respectively. 
The NE-EA uses a (20+2) population with 1000 iterations for the behavior search and (8+2), 500 for the surrogate.  
The critic network is a fully-connected feed-forward ANN with two layers and 128/64 tanh units, trained for 1000 steps in each iteration. 
The prototype code and all experimental results are available in an online repository: \url{https://github.com/jstork/BNET-GECCO21}.

\subsection{Open AI Gym Benchmark Setup}
For comparing the performance of BNET against different baselines, we chose two basic Open AI Gym standard benchmarks. They were explicitly chosen because of their different characteristics.
Moreover, both should be solvable in less than 20,000 steps. 
They present a baseline for a real-world scenario, where typically only a low number of steps is possible due to the high resource cost.

\textbf{Cartpole-v1} is a standard benchmark, where the goal is to balance a pole placed on a cart. The environment has four observable variables ($x$ position, $x$ velocity, pole angle, angular velocity) and two discrete actions (drive left or right). The target is to keep the pole upright in a slight angle range for an average of 195 steps over 100 consecutive episodes. If the pole fails to balance (i.e., the angle reaches a threshold) or the cart drives out of a certain x-range, an episode is stopped. For each step, the agent receives a reward of $+1$.

In \textbf{MountainCar-v0}, a car is located between two mountains and has to drive to the top of one of them. 
As the direct acceleration is not high enough, it has to build up momentum by alternatively driving up and down the mountains. 
The environment has two observable variables (x-position, velocity) and three actions (accelerate left/right, do nothing). 
The target is to drive to a goal position on the proper mountain in less than average 110 time-steps. Each step is rewarded by -1, and the environment is stopped if either the step limit (200) or the goal is reached. The environment requires considerable exploration to find a solution to reach the goal point. 
If the exploration is unsuccessful, it remains with a -200 reward in each episode and gains no valuable experience. 
This flat reward landscape renders the environment as difficult to solve in a small number of steps. 
For both environments, the starting state of the pole or car is randomly set in a small range, leading to different initial scenarios for each episode. 
Therefore, each setup was repeated at least 20 times with random initial seeds. 
The run was stopped if a found policy reached the required average target, which was evaluated in an extra function to save unnecessary computation time. 
The fitness or experiences of these stopping criteria evaluations were not utilized in any other form (e.g., for the algorithm itself).

\begin{figure}[t]
\centering
\includegraphics[width=0.47\textwidth]{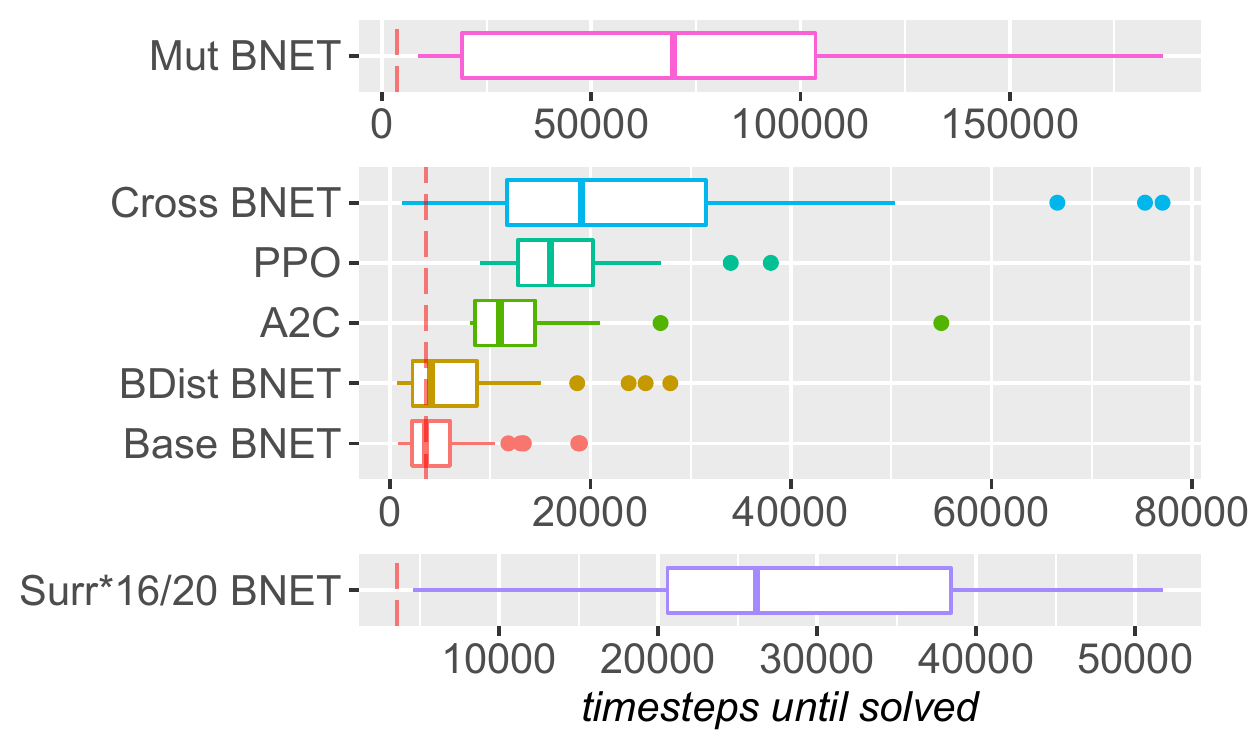}
\caption{Results of the CartPole-v0 environment. Please mind the different scales for the surrogate and mutation variant. The surrogate variant was only able to finish in 16 out of 20 runs. The best results are achieved by generating all candidates (Base median=3576, red dashed line) or only the behavior distance candidate (BDist with median=4111).
}
\label{fig:cartpole}
\end{figure}

The first benchmarks include setup variants of BNET, where selected candidates were generated in each iteration. 
The elitist was always kept and repeated (for the robust evaluation). The setups are: \emph{Base} (all proposed modules are active), \emph{BDist} (behavior distance), \emph{Cross} (cross-entropy), \emph{Surr} (SMBO) and \emph{Mut} (champion mutation). 
For each variant, the population consists of the champion and a single candidate per active module (e.g., BDist has two candidates per iteration), and the maximum number of episodes was fixed to 1000, except for the \emph{Mut} variant, which served as an additional internal baseline and was run until the environments were solved. 
In MountainCar-v0, we always kept the mutation candidate in the population. Further, we added additional random exploration (30\%) to its policy. Random exploration was also added to the environment evaluations of the initial candidates. 
All remaining policies are always evaluated deterministically.
Each run starts with five initial, random candidates, while the elitist experience set contains a maximum of ten archived episode results. 

\subsection{OpenAI Benchmark Results}

\begin{figure}[t]
\centering
\includegraphics[width=0.47\textwidth]{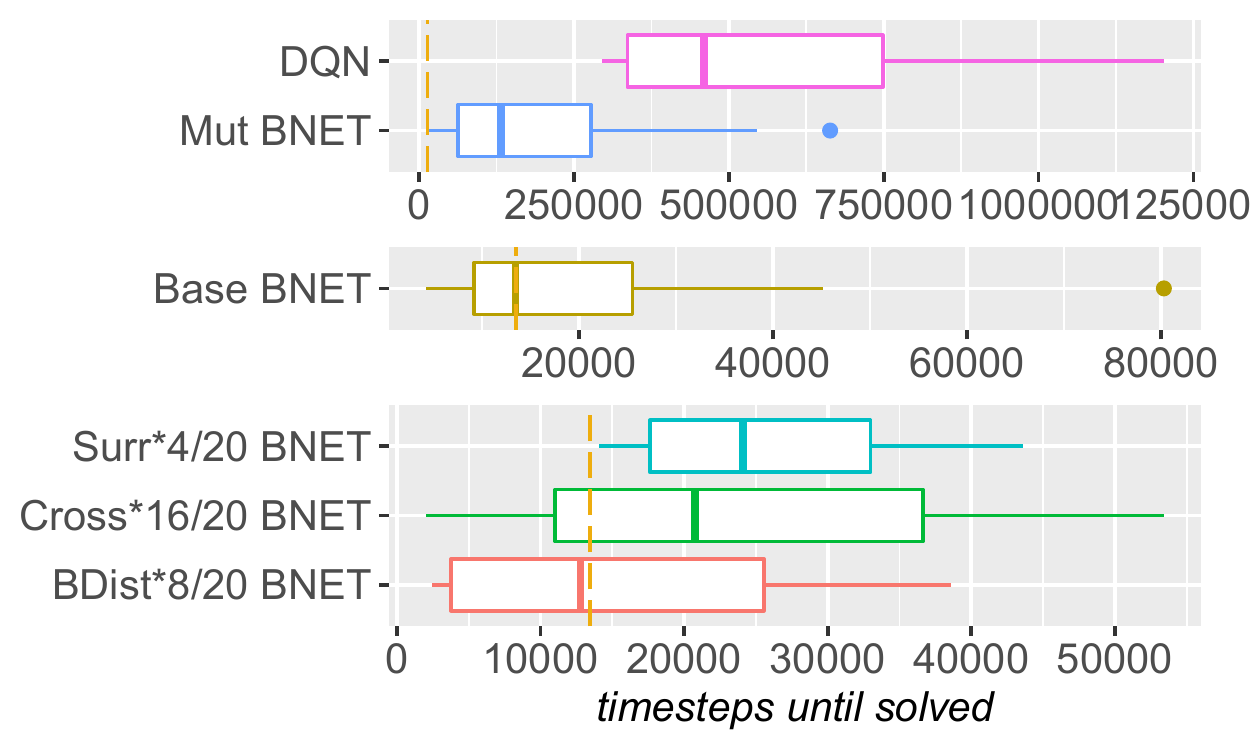}
\caption{MountainCar-v0 results. Please mind the different scales. The Base variant was able to solve the environment in all cases (median=13457, gold dashed line). The results of the unfinished runs are not comparable. In their case, the attached number of finished runs is meaningful. 
}
\label{fig:mcarRes}
\end{figure}

\begin{figure}[t]
\centering
\includegraphics[width=0.4\textwidth]{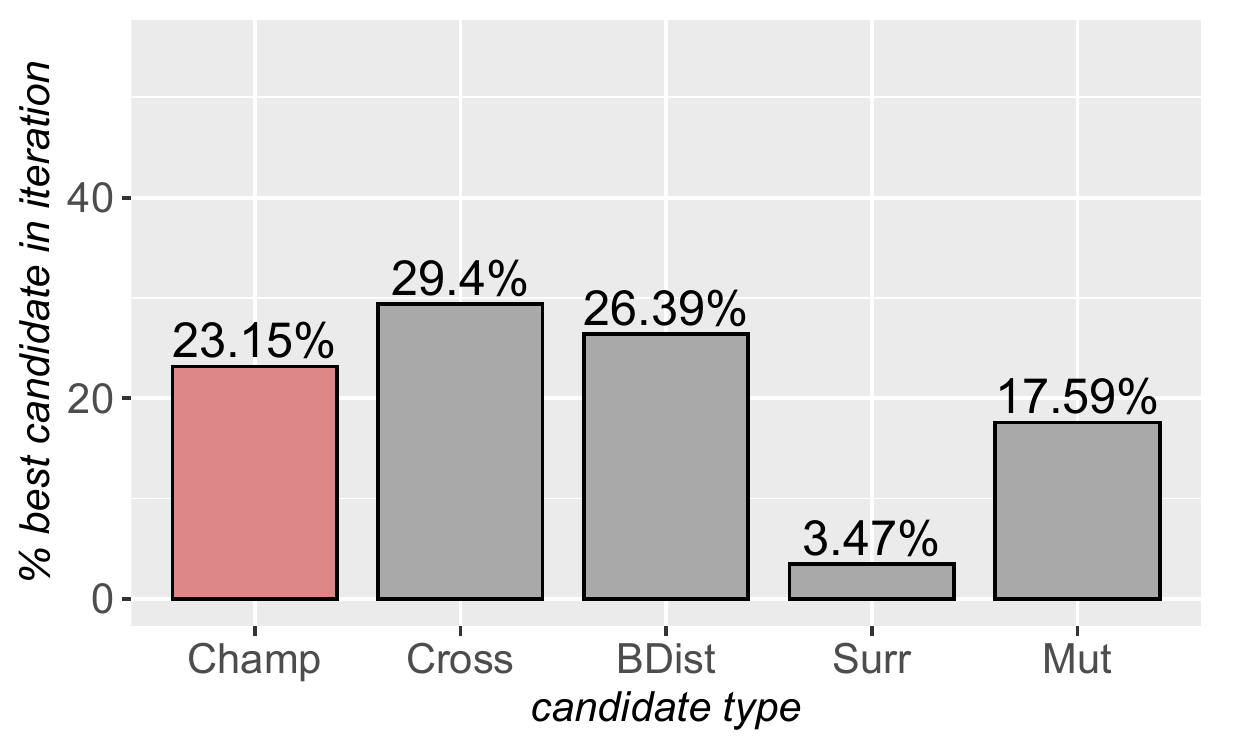}
\caption{Frequency proportion of each best performing candidate in each iteration of Base BNET for CartPole-v0.
}
\label{fig:cartpoleType}
\end{figure}

\textbf{(Cartpole-v0)} Fig. \ref{fig:cartpole} illustrates our results from the CartPole-V0 environment. 
For the plot, each algorithm was repeated 20 times, except for the \emph{Base} and \emph{BDist} variants, which were repeated 50 times (in order to make more minor differences visible). The surrogate-only variant of BNET only succeeded in 16 out of 20 runs to find a solution in 1000 iterations. Overall, the \emph{BDist} variant only generating the behavior-distance optimized candidate (and keeping the robust champion), and the \emph{Base} version using all proposed search methods were the most successful. 
As expected, the use of only direct mutation performed slower than all other variants. 
The overall result was surprising for us, as we expected that one variant beside \emph{Base} would be the best performing, as it includes a relatively large sampling overhead by the larger population. 
However, the algorithm seems to leverage from a diverse set of candidates, and each search variant seems to contribute to the overall algorithm's performance.
To test this assumption, we tracked the candidate type with the best fitness (mean over 100 iterations) in each iteration of the BNET \emph{Base} setup. Fig. \ref{fig:cartpoleType} shows the results.
The underlying data is from the 50 repeats of the BNET \emph{Base} runs, with 432 iterations. 


The plot shows two insights: first, in 77\% of the iterations, one of the generated candidates was superior to the stored champion, this implies a reasonable learning rate; second, all candidates of BNET contribute to the performance, where only the surrogate selected candidate performs significantly worse. 
The surrogate candidate's inferiority could be due to poor parametrization or due to the low number of evaluations, which might not be sufficient to create a proper surrogate model for the complex search space. 
Interestingly, the direct mutation also generated the best candidates and thus added significantly to the overall performance. 

\textbf{(MountainCar-v0)} Due to the challenging nature of this problem, we were not able to find working setups for our baseline algorithms A2C and PPO. All tested parametrization showed no learning effect and got stuck at a reward level of -200, even considering significantly large timestep budgets. 
We thus tested an additional algorithm, \emph{Deep Q Networks} (DQN)~\cite{mnih2013playing} in available setups (double-DQN, dueling-DQN, prioritized experience replay) and were able to solve the environment using prioritized experience replay~\cite{schaul2015prioritized} and high exploration constants. However, DQN still took a vast number of steps to solve the environment and performs even inferior to the BNET mutation variant. 
Fig. \ref{fig:mcarRes} displays the MountainCar-v0 results.
As visible, the BNET Base variant is dominating this benchmark and remains the only variant that solves the environment in the 100k step limit. Still, the other algorithms' result is quite interesting, as they tend to either solve the environment in a small number of steps or seem to get completely stuck. 
As Fig. \ref{fig:mcarRes} illustrates, the percentage of successful candidates in the Base variant per iteration is 47\%, much less compared to CartPole-v0, with a clear lead of cross-entropy and mutation. 
The BDist optimization clearly falls behind for this environment, visible in both Fig \ref{fig:mcarRes} and \ref{fig:mcarRes}. 
We assume this issue is related to the problematic value estimation due to the flat reward landscape of MountainCar-v0.
Again, the surrogate search does not perform as desired, which is also caused by the very flat fitness landscape at the beginning, which does not include much valuable information. 

\subsection{Real-World Robot Maze Setup}

Our robot maze problem was explicitly designed to represent a costly real-world demonstrator to test RL algorithms on different setups. 
We chose a classic maze problem to track and observe an agent's progress and performance efficiently.

\begin{figure}[b]
\centering
\includegraphics[width=0.3\textwidth]{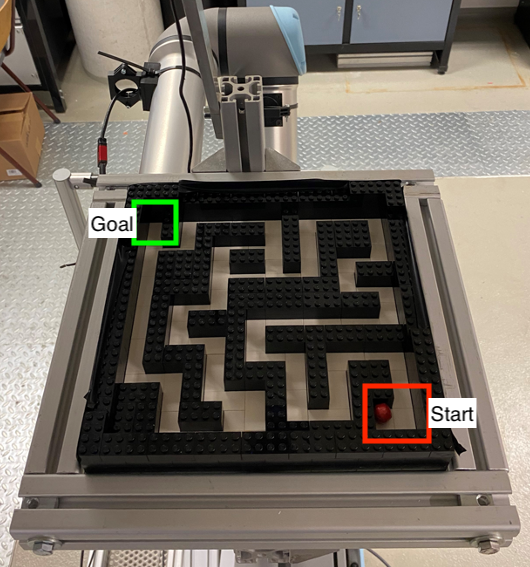}
\caption{Robot maze test environment mounted on a universal robots UR-10 collaborative robot.}
\label{fig:robotmaze}
\end{figure}

The maze consists of a lego brick base plate with 250 x 250 mm, 4x2 black brick walls, and 4x2 white tiles floor, covered by an acrylic glass cover, where a camera is mounted on top. 
The camera is used to track the position of the red marble in the maze. 
The system is mounted on a universal robot UR10e 6-axis robotic arm, allowing to move the maze in all directions.
The setup is displayed in Fig. \ref{fig:robotmaze}. The target is to move the marble to the upper left position from the starting point.
A central challenge in designing real-world problems is learning without manual user interaction (i.e., resetting a robot position). 
Our demonstrator allows the automatic resetting by flipping the complete maze and navigating the marble on the glass window to the start. This reset allows episode-to-episode learning and further remote control of the environment without any presence in the lab.
The demonstrator is adaptable, i.e., the maze can be redesigned, and the action and observation space can be adapted to discrete or continuous values.

\begin{figure}[t]
\centering
\includegraphics[width=0.4\textwidth]{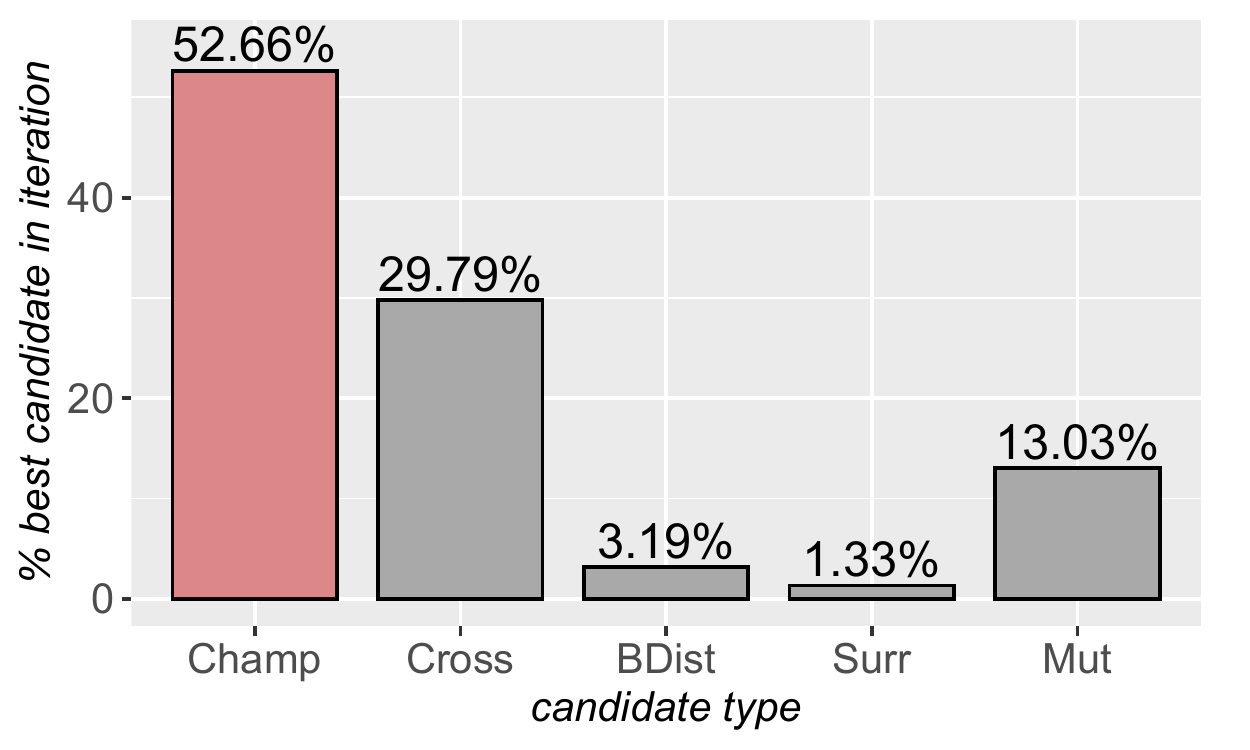}
\caption{Frequency proportion of each best performing candidate in each iteration of Base BNET for MountainCar-v0.
}
\label{fig:mcarRes}
\end{figure}

\label{sec:realworld}

For this work, we restricted the action space to the discrete four cardinal directions and defined designated robot movements, which tilt the maze by an angle of 24,6° and then move back to its base position. 
Each action takes about 10 seconds and lets the marble roll in a straight line. 
For the illustrated maze setup, only 23 correct actions are required to reach the target area. 
The observation space is set to a discrete matrix of 15x15, equal to the maze size, where the marble's current position is set to one, else zero.
The reward function forces exploration of the maze by rewarding the agent with 0.1 if he drives the marble to a prior unseen position. 
If run against a wall, it is penalized by -0.75 and by -0.25 if moved to an already discovered position. 
Reaching the goal is worth +10. 
The setup is reset if the goal, a number of 75 steps, or a cumulative reward of -12.5 is reached.
The robot maze environment should be fast to learn, as, for most positions, only a single action is correct.
However, it requires a perfect mixture of exploration and exploitation to find and learn the correct actions sequentially, as the probability of getting stuck is high, and only a small number of steps is possible in each episode. 
Moreover, the environment fitness is noisy, as sometimes the marble rolls to a difficult position (i.e., edges of a brick), or the camera position detections are incorrect. 

\begin{figure}[t]
\centering
\includegraphics[width=0.4\textwidth]{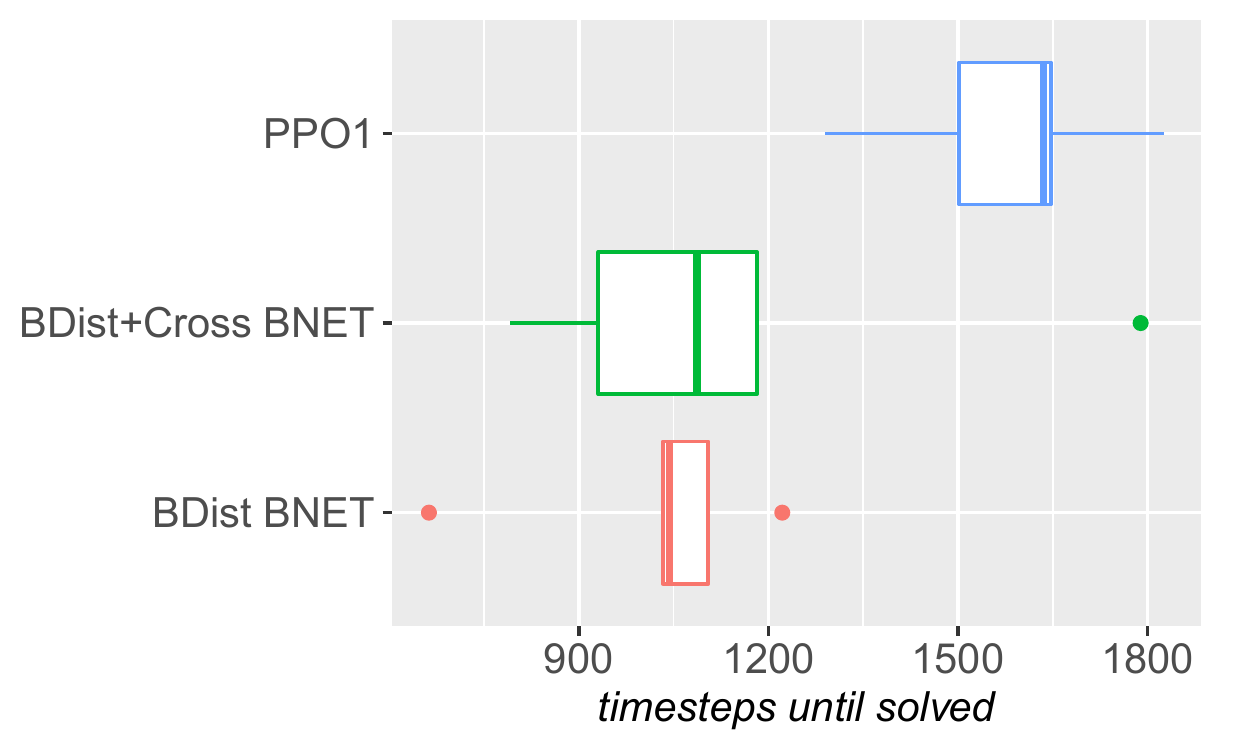}
\caption{Maze results. Steps until the correct way is learned; The target is consequently reached afterwards. All tested algorithms learn fast. Means (top-down) = 1580, 1156, 1013.}
\label{fig:robotres}
\end{figure}

Due to the natural time constraints (each run took 8-24h) and several trials to set up the environment, we could not run an extensive experimental design for the benchmark. Thus, we restricted the final results to the most promising ones for this setup: the behavior distance and cross-entropy versions. Each variant was run five times. 
Regarding the algorithm setup, we set the fitness value in BNET to the number of correctly performed steps. 
The behavior and cross-entropy distances were weighted by the direct reward instead of a critic and advantage estimation. 
The purpose-built reward function already delivers correct action value information, and no additional value critic is required.
As a baseline, we utilize PPO that showed very stable results in preliminary runs. We set gamma=0.5 due to the apparent correlation between correct actions and rewards and 50 steps per actorbatch for frequent learning. 

Fig. \ref{fig:robotres} illustrates the results. All pre-selected algorithms were quite fast in solving the problem, with the \emph{BDist} and \emph{BDist+Cross} versions performing best. In Fig. \ref{fig:robotdot} we visualize the learning progress of the \emph{BDist+Cross} runs. The algorithm advances quite fast and even learns to take more reward-giving actions (>23), assumingly by taking extra detours in the maze. This behavior is similar for PPO and caused by the definition of the rewarding of previous unvisited positions. 
However, the underlying NE-EA with the behavior-based losses demonstrates to be capable of fitting the CGP-ANN policies excellently to the collected experience, even given the sizeable neuroevolutionary search space, and without the use of gradients. 

\section{Conclusion and Outlook}
\label{sec:conclusion}
In this work, we investigated methods to combine neuroevolutionary search with standard value-based RL methods. 
The target was to leverage from gathered experiences and to create a sample-efficient RL algorithm applicable to real-world problems.
We evolve CGP-ANNs as our agents' policies and search for the best network topologies and optimal weights simultaneously.  
We defined a hybrid, population-based algorithm, called BNET, which utilizes different methods to generate new candidates: direct NE-based mutation, behavior optimization using gathered experience, and finally, behavior-surrogate-based learning. 
We discovered that the behavior-based search significantly supports the performance. Combining all methods in a population with shared experience and fitness pools leads to excellent sample efficiency and extraordinary explorative abilities. 
Moreover, even elementary direct neuroevolutionary mutation steps can contribute significantly to the overall algorithm's performance. 
As our real-world experiment demonstrates, the defined behavior-loss functions seem well suited to optimize complex networks with changing topologies if the actions' value is estimated correctly. 
Furthermore, the robust selection with an adaptive re-evaluation of candidates significantly improves our learning progress' stability, as shown in the experiments with a real-world setup. 
They prove the ability to learn fast and adapt the CGP-ANN policies by solely relying on a gradient-free evolutionary algorithm for optimization. 
In future work, we want to tackle several open issues:


\begin{figure}[t]
\centering
\includegraphics[width=0.4\textwidth]{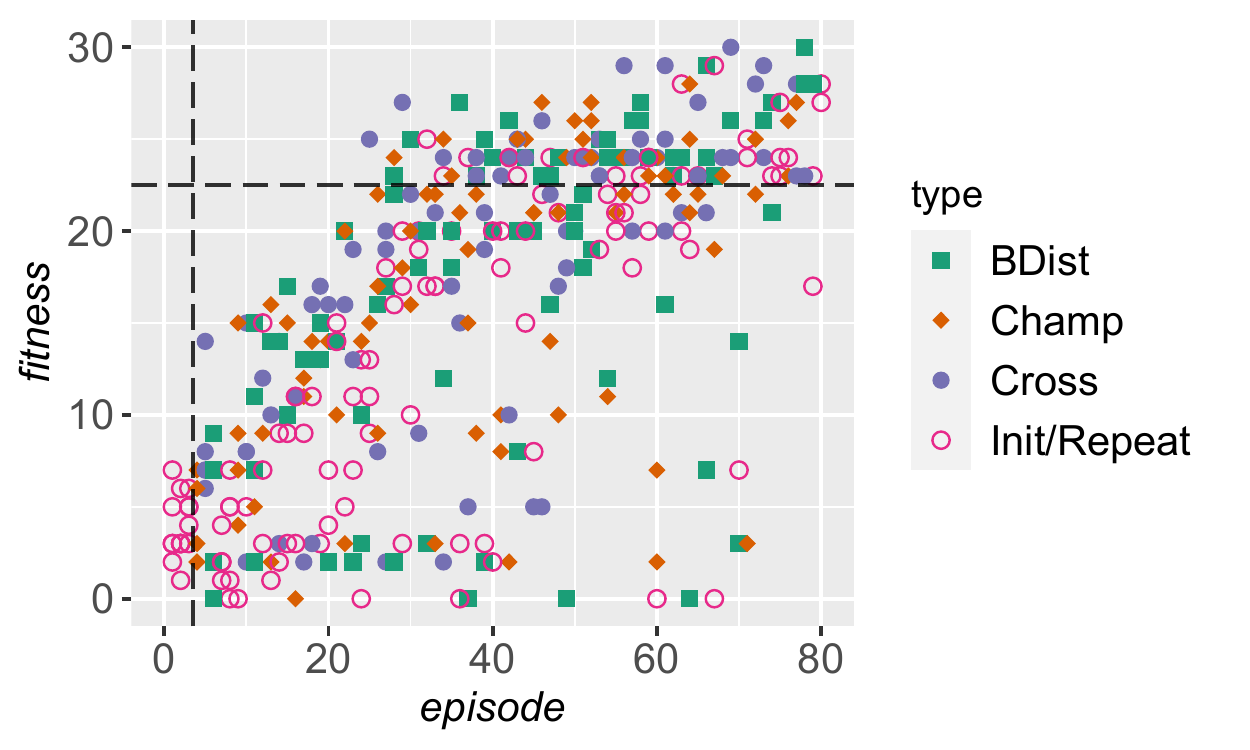}
\caption{Learning progress of all runs of BDist+Cross BNET on robotmaze. Considered solved at fitness>=23 (horizontal line). BNET shows fast and stable learning process.}
\label{fig:robotdot}
\end{figure}
 
\textbf{Framework}: We presented an implementation prototype of BNET. 
The current version is rather slow due to a single-thread implementation in R. 
The underlying ideas need to be transferred to a faster and computationally more efficient implementation.

\textbf{Analysis / Tuning}: The algorithm structure and parameters need to be profoundly analyzed, understood, and optimized.

\textbf{Offline initialization}: Real-world problems can benefit considerably from experience from prior runs or human demonstrations. We want to implement and test offline-initialization methods. 
 
\textbf{Environments}: We focused on discrete action spaces. Extending benchmarks to continuous action spaces would be interesting. 

\textbf{Modified Real-World Experiment}: Our real-world experiment can be adapted to create more challenging RL problems.

\bibliographystyle{ACM-Reference-Format}
\bibliography{Stor20c} 


\begin{thebibliography}{35}


\ifx \showCODEN    \undefined \def \showCODEN     #1{\unskip}     \fi
\ifx \showDOI      \undefined \def \showDOI       #1{#1}\fi
\ifx \showISBNx    \undefined \def \showISBNx     #1{\unskip}     \fi
\ifx \showISBNxiii \undefined \def \showISBNxiii  #1{\unskip}     \fi
\ifx \showISSN     \undefined \def \showISSN      #1{\unskip}     \fi
\ifx \showLCCN     \undefined \def \showLCCN      #1{\unskip}     \fi
\ifx \shownote     \undefined \def \shownote      #1{#1}          \fi
\ifx \showarticletitle \undefined \def \showarticletitle #1{#1}   \fi
\ifx \showURL      \undefined \def \showURL       {\relax}        \fi
\providecommand\bibfield[2]{#2}
\providecommand\bibinfo[2]{#2}
\providecommand\natexlab[1]{#1}
\providecommand\showeprint[2][]{arXiv:#2}

\bibitem[\protect\citeauthoryear{Abadi et~al\mbox{.}}{Abadi
  et~al\mbox{.}}{2015}]%
        {tensorflow2015-whitepaper}
\bibfield{author}{\bibinfo{person}{Mart\'{\i}n Abadi} {et~al\mbox{.}}}
  \bibinfo{year}{2015}\natexlab{}.
\newblock \bibinfo{title}{{TensorFlow}: Large-Scale Machine Learning on
  Heterogeneous Systems}.
\newblock
\newblock
\urldef\tempurl%
\url{https://www.tensorflow.org/}
\showURL{%
\tempurl}
\newblock
\shownote{Software available from tensorflow.org.}


\bibitem[\protect\citeauthoryear{Brockman, Cheung, Pettersson, Schneider,
  Schulman, Tang, and Zaremba}{Brockman et~al\mbox{.}}{2016}]%
        {brockman2016openai}
\bibfield{author}{\bibinfo{person}{Greg Brockman}, \bibinfo{person}{Vicki
  Cheung}, \bibinfo{person}{Ludwig Pettersson}, \bibinfo{person}{Jonas
  Schneider}, \bibinfo{person}{John Schulman}, \bibinfo{person}{Jie Tang},
  {and} \bibinfo{person}{Wojciech Zaremba}.} \bibinfo{year}{2016}\natexlab{}.
\newblock \showarticletitle{Openai gym}.
\newblock \bibinfo{journal}{\emph{arXiv preprint arXiv:1606.01540}}
  (\bibinfo{year}{2016}).
\newblock


\bibitem[\protect\citeauthoryear{Chollet et~al\mbox{.}}{Chollet
  et~al\mbox{.}}{2015}]%
        {chollet2015keras}
\bibfield{author}{\bibinfo{person}{Fran\c{c}ois Chollet} {et~al\mbox{.}}}
  \bibinfo{year}{2015}\natexlab{}.
\newblock \bibinfo{title}{Keras}.
\newblock \bibinfo{howpublished}{\url{https://keras.io}}.
\newblock


\bibitem[\protect\citeauthoryear{Eiben, Smith, et~al\mbox{.}}{Eiben
  et~al\mbox{.}}{2003}]%
        {eiben2003introduction}
\bibfield{author}{\bibinfo{person}{Agoston~E Eiben}, \bibinfo{person}{James~E
  Smith}, {et~al\mbox{.}}} \bibinfo{year}{2003}\natexlab{}.
\newblock \bibinfo{booktitle}{\emph{Introduction to evolutionary computing}}.
  Vol.~\bibinfo{volume}{53}.
\newblock \bibinfo{publisher}{Springer}.
\newblock


\bibitem[\protect\citeauthoryear{Forrester, Sobester, and Keane}{Forrester
  et~al\mbox{.}}{2008}]%
        {Forrester2008a}
\bibfield{author}{\bibinfo{person}{Alexander Forrester},
  \bibinfo{person}{Andras Sobester}, {and} \bibinfo{person}{Andy Keane}.}
  \bibinfo{year}{2008}\natexlab{}.
\newblock \bibinfo{booktitle}{\emph{Engineering Design via Surrogate
  Modelling}}.
\newblock \bibinfo{publisher}{Wiley}.
\newblock


\bibitem[\protect\citeauthoryear{Franke, K{\"o}hler, Awad, and Hutter}{Franke
  et~al\mbox{.}}{2019}]%
        {franke2019neural}
\bibfield{author}{\bibinfo{person}{J{\"o}rg~KH Franke}, \bibinfo{person}{Gregor
  K{\"o}hler}, \bibinfo{person}{Noor Awad}, {and} \bibinfo{person}{Frank
  Hutter}.} \bibinfo{year}{2019}\natexlab{}.
\newblock \showarticletitle{Neural Architecture Evolution in Deep Reinforcement
  Learning for Continuous Control}.
\newblock \bibinfo{journal}{\emph{arXiv preprint arXiv:1910.12824}}
  (\bibinfo{year}{2019}).
\newblock


\bibitem[\protect\citeauthoryear{Gablonsky and Kelley}{Gablonsky and
  Kelley}{2001}]%
        {gablonsky2001locally}
\bibfield{author}{\bibinfo{person}{Joerg~M Gablonsky} {and}
  \bibinfo{person}{Carl~T Kelley}.} \bibinfo{year}{2001}\natexlab{}.
\newblock \showarticletitle{A locally-biased form of the DIRECT algorithm}.
\newblock \bibinfo{journal}{\emph{Journal of Global Optimization}}
  \bibinfo{volume}{21}, \bibinfo{number}{1} (\bibinfo{year}{2001}),
  \bibinfo{pages}{27--37}.
\newblock


\bibitem[\protect\citeauthoryear{Gaier, Asteroth, and Mouret}{Gaier
  et~al\mbox{.}}{2018}]%
        {gaier2018data}
\bibfield{author}{\bibinfo{person}{Adam Gaier}, \bibinfo{person}{Alexander
  Asteroth}, {and} \bibinfo{person}{Jean-Baptiste Mouret}.}
  \bibinfo{year}{2018}\natexlab{}.
\newblock \showarticletitle{Data-efficient neuroevolution with kernel-based
  surrogate models}. In \bibinfo{booktitle}{\emph{Proceedings of the genetic
  and evolutionary computation conference}}. \bibinfo{pages}{85--92}.
\newblock


\bibitem[\protect\citeauthoryear{Hill, Raffin, Ernestus, Gleave, Kanervisto,
  Traore, Dhariwal, Hesse, Klimov, Nichol, Plappert, Radford, Schulman, Sidor,
  and Wu}{Hill et~al\mbox{.}}{2018}]%
        {stable-baselines}
\bibfield{author}{\bibinfo{person}{Ashley Hill}, \bibinfo{person}{Antonin
  Raffin}, \bibinfo{person}{Maximilian Ernestus}, \bibinfo{person}{Adam
  Gleave}, \bibinfo{person}{Anssi Kanervisto}, \bibinfo{person}{Rene Traore},
  \bibinfo{person}{Prafulla Dhariwal}, \bibinfo{person}{Christopher Hesse},
  \bibinfo{person}{Oleg Klimov}, \bibinfo{person}{Alex Nichol},
  \bibinfo{person}{Matthias Plappert}, \bibinfo{person}{Alec Radford},
  \bibinfo{person}{John Schulman}, \bibinfo{person}{Szymon Sidor}, {and}
  \bibinfo{person}{Yuhuai Wu}.} \bibinfo{year}{2018}\natexlab{}.
\newblock \bibinfo{title}{Stable Baselines}.
\newblock
  \bibinfo{howpublished}{\url{https://github.com/hill-a/stable-baselines}}.
\newblock


\bibitem[\protect\citeauthoryear{Izzo, Biscani, and Mereta}{Izzo
  et~al\mbox{.}}{2017}]%
        {izzo2017differentiable}
\bibfield{author}{\bibinfo{person}{Dario Izzo}, \bibinfo{person}{Francesco
  Biscani}, {and} \bibinfo{person}{Alessio Mereta}.}
  \bibinfo{year}{2017}\natexlab{}.
\newblock \showarticletitle{Differentiable genetic programming}. In
  \bibinfo{booktitle}{\emph{European conference on genetic programming}}.
  Springer, \bibinfo{pages}{35--51}.
\newblock


\bibitem[\protect\citeauthoryear{Jaderberg, Dalibard, Osindero, Czarnecki,
  Donahue, Razavi, Vinyals, Green, Dunning, Simonyan, et~al\mbox{.}}{Jaderberg
  et~al\mbox{.}}{2017}]%
        {jaderberg2017population}
\bibfield{author}{\bibinfo{person}{Max Jaderberg}, \bibinfo{person}{Valentin
  Dalibard}, \bibinfo{person}{Simon Osindero}, \bibinfo{person}{Wojciech~M
  Czarnecki}, \bibinfo{person}{Jeff Donahue}, \bibinfo{person}{Ali Razavi},
  \bibinfo{person}{Oriol Vinyals}, \bibinfo{person}{Tim Green},
  \bibinfo{person}{Iain Dunning}, \bibinfo{person}{Karen Simonyan},
  {et~al\mbox{.}}} \bibinfo{year}{2017}\natexlab{}.
\newblock \showarticletitle{Population based training of neural networks}.
\newblock \bibinfo{journal}{\emph{arXiv preprint arXiv:1711.09846}}
  (\bibinfo{year}{2017}).
\newblock


\bibitem[\protect\citeauthoryear{Jones, Schonlau, and Welch}{Jones
  et~al\mbox{.}}{1998}]%
        {Jones1998}
\bibfield{author}{\bibinfo{person}{Donald~R. Jones}, \bibinfo{person}{Matthias
  Schonlau}, {and} \bibinfo{person}{William~J. Welch}.}
  \bibinfo{year}{1998}\natexlab{}.
\newblock \showarticletitle{Efficient global optimization of expensive
  black-box functions}.
\newblock \bibinfo{journal}{\emph{Journal of Global Optimization}}
  \bibinfo{volume}{13}, \bibinfo{number}{4} (\bibinfo{year}{1998}),
  \bibinfo{pages}{455--492}.
\newblock


\bibitem[\protect\citeauthoryear{Jung, Park, and Sung}{Jung
  et~al\mbox{.}}{2020}]%
        {jung2020population}
\bibfield{author}{\bibinfo{person}{Whiyoung Jung}, \bibinfo{person}{Giseung
  Park}, {and} \bibinfo{person}{Youngchul Sung}.}
  \bibinfo{year}{2020}\natexlab{}.
\newblock \showarticletitle{Population-guided parallel policy search for
  reinforcement learning}.
\newblock \bibinfo{journal}{\emph{arXiv preprint arXiv:2001.02907}}
  (\bibinfo{year}{2020}).
\newblock


\bibitem[\protect\citeauthoryear{Khadka, Majumdar, Nassar, Dwiel, Tumer, Miret,
  Liu, and Tumer}{Khadka et~al\mbox{.}}{2019}]%
        {khadka2019collaborative}
\bibfield{author}{\bibinfo{person}{Shauharda Khadka}, \bibinfo{person}{Somdeb
  Majumdar}, \bibinfo{person}{Tarek Nassar}, \bibinfo{person}{Zach Dwiel},
  \bibinfo{person}{Evren Tumer}, \bibinfo{person}{Santiago Miret},
  \bibinfo{person}{Yinyin Liu}, {and} \bibinfo{person}{Kagan Tumer}.}
  \bibinfo{year}{2019}\natexlab{}.
\newblock \showarticletitle{Collaborative evolutionary reinforcement learning}.
  In \bibinfo{booktitle}{\emph{International Conference on Machine Learning}}.
  PMLR, \bibinfo{pages}{3341--3350}.
\newblock


\bibitem[\protect\citeauthoryear{Khadka and Tumer}{Khadka and Tumer}{2018}]%
        {khadka2018evolution}
\bibfield{author}{\bibinfo{person}{Shauharda Khadka} {and}
  \bibinfo{person}{Kagan Tumer}.} \bibinfo{year}{2018}\natexlab{}.
\newblock \showarticletitle{Evolution-guided policy gradient in reinforcement
  learning}. In \bibinfo{booktitle}{\emph{Proceedings of the 32nd International
  Conference on Neural Information Processing Systems}}.
  \bibinfo{pages}{1196--1208}.
\newblock


\bibitem[\protect\citeauthoryear{Miller and Thomson}{Miller and
  Thomson}{2000}]%
        {miller2000cartesian}
\bibfield{author}{\bibinfo{person}{Julian~F Miller} {and}
  \bibinfo{person}{Peter Thomson}.} \bibinfo{year}{2000}\natexlab{}.
\newblock \showarticletitle{Cartesian genetic programming}. In
  \bibinfo{booktitle}{\emph{European Conference on Genetic Programming}}.
  Springer, \bibinfo{pages}{121--132}.
\newblock


\bibitem[\protect\citeauthoryear{Mnih, Badia, Mirza, Graves, Lillicrap, Harley,
  Silver, and Kavukcuoglu}{Mnih et~al\mbox{.}}{2016}]%
        {mnih2016asynchronous}
\bibfield{author}{\bibinfo{person}{Volodymyr Mnih},
  \bibinfo{person}{Adria~Puigdomenech Badia}, \bibinfo{person}{Mehdi Mirza},
  \bibinfo{person}{Alex Graves}, \bibinfo{person}{Timothy Lillicrap},
  \bibinfo{person}{Tim Harley}, \bibinfo{person}{David Silver}, {and}
  \bibinfo{person}{Koray Kavukcuoglu}.} \bibinfo{year}{2016}\natexlab{}.
\newblock \showarticletitle{Asynchronous methods for deep reinforcement
  learning}. In \bibinfo{booktitle}{\emph{International conference on machine
  learning}}. PMLR, \bibinfo{pages}{1928--1937}.
\newblock


\bibitem[\protect\citeauthoryear{Mnih, Kavukcuoglu, Silver, Graves, Antonoglou,
  Wierstra, and Riedmiller}{Mnih et~al\mbox{.}}{2013}]%
        {mnih2013playing}
\bibfield{author}{\bibinfo{person}{Volodymyr Mnih}, \bibinfo{person}{Koray
  Kavukcuoglu}, \bibinfo{person}{David Silver}, \bibinfo{person}{Alex Graves},
  \bibinfo{person}{Ioannis Antonoglou}, \bibinfo{person}{Daan Wierstra}, {and}
  \bibinfo{person}{Martin Riedmiller}.} \bibinfo{year}{2013}\natexlab{}.
\newblock \showarticletitle{Playing atari with deep reinforcement learning}.
\newblock \bibinfo{journal}{\emph{arXiv preprint arXiv:1312.5602}}
  (\bibinfo{year}{2013}).
\newblock


\bibitem[\protect\citeauthoryear{Mockus, Tiesis, and Zilinskas}{Mockus
  et~al\mbox{.}}{1978}]%
        {Mockus1978}
\bibfield{author}{\bibinfo{person}{Jonas Mockus}, \bibinfo{person}{Vytautas
  Tiesis}, {and} \bibinfo{person}{Antanas Zilinskas}.}
  \bibinfo{year}{1978}\natexlab{}.
\newblock \bibinfo{booktitle}{\emph{Towards Global Optimization 2}}.
\newblock \bibinfo{publisher}{North-Holland}, Chapter The application of
  Bayesian methods for seeking the extremum, \bibinfo{pages}{117--129}.
\newblock


\bibitem[\protect\citeauthoryear{Moraglio and Kattan}{Moraglio and
  Kattan}{2011}]%
        {Moraglio2011}
\bibfield{author}{\bibinfo{person}{Alberto Moraglio} {and}
  \bibinfo{person}{Ahmed Kattan}.} \bibinfo{year}{2011}\natexlab{}.
\newblock \showarticletitle{Geometric Generalisation of Surrogate Model Based
  Optimisation to Combinatorial Spaces}. In
  \bibinfo{booktitle}{\emph{Proceedings of the 11th European Conference on
  Evolutionary Computation in Combinatorial Optimization}} (Torino, Italy)
  \emph{(\bibinfo{series}{EvoCOP'11})}. \bibinfo{publisher}{Springer},
  \bibinfo{pages}{142--154}.
\newblock


\bibitem[\protect\citeauthoryear{Parker-Holder, Pacchiano, Choromanski, and
  Roberts}{Parker-Holder et~al\mbox{.}}{2020}]%
        {parker2020effective}
\bibfield{author}{\bibinfo{person}{Jack Parker-Holder}, \bibinfo{person}{Aldo
  Pacchiano}, \bibinfo{person}{Krzysztof Choromanski}, {and}
  \bibinfo{person}{Stephen Roberts}.} \bibinfo{year}{2020}\natexlab{}.
\newblock \showarticletitle{Effective diversity in population-based
  reinforcement learning}.
\newblock \bibinfo{journal}{\emph{arXiv preprint arXiv:2002.00632}}
  (\bibinfo{year}{2020}).
\newblock


\bibitem[\protect\citeauthoryear{Rehbach, Zaefferer, Naujoks, and
  Bartz-Beielstein}{Rehbach et~al\mbox{.}}{2020}]%
        {rehbach2020expected}
\bibfield{author}{\bibinfo{person}{Frederik Rehbach}, \bibinfo{person}{Martin
  Zaefferer}, \bibinfo{person}{Boris Naujoks}, {and} \bibinfo{person}{Thomas
  Bartz-Beielstein}.} \bibinfo{year}{2020}\natexlab{}.
\newblock \showarticletitle{Expected improvement versus predicted value in
  surrogate-based optimization}. In \bibinfo{booktitle}{\emph{Proceedings of
  the 2020 Genetic and Evolutionary Computation Conference}}.
  \bibinfo{publisher}{ACM}, \bibinfo{pages}{868--876}.
\newblock


\bibitem[\protect\citeauthoryear{Salimans, Ho, Chen, Sidor, and
  Sutskever}{Salimans et~al\mbox{.}}{2017}]%
        {salimans2017evolution}
\bibfield{author}{\bibinfo{person}{Tim Salimans}, \bibinfo{person}{Jonathan
  Ho}, \bibinfo{person}{Xi Chen}, \bibinfo{person}{Szymon Sidor}, {and}
  \bibinfo{person}{Ilya Sutskever}.} \bibinfo{year}{2017}\natexlab{}.
\newblock \showarticletitle{Evolution strategies as a scalable alternative to
  reinforcement learning}.
\newblock \bibinfo{journal}{\emph{arXiv preprint arXiv:1703.03864}}
  (\bibinfo{year}{2017}).
\newblock


\bibitem[\protect\citeauthoryear{Schaul, Quan, Antonoglou, and Silver}{Schaul
  et~al\mbox{.}}{2015}]%
        {schaul2015prioritized}
\bibfield{author}{\bibinfo{person}{Tom Schaul}, \bibinfo{person}{John Quan},
  \bibinfo{person}{Ioannis Antonoglou}, {and} \bibinfo{person}{David Silver}.}
  \bibinfo{year}{2015}\natexlab{}.
\newblock \showarticletitle{Prioritized experience replay}.
\newblock \bibinfo{journal}{\emph{arXiv preprint arXiv:1511.05952}}
  (\bibinfo{year}{2015}).
\newblock


\bibitem[\protect\citeauthoryear{Schmitt, Hessel, and Simonyan}{Schmitt
  et~al\mbox{.}}{2020}]%
        {schmitt2020off}
\bibfield{author}{\bibinfo{person}{Simon Schmitt}, \bibinfo{person}{Matteo
  Hessel}, {and} \bibinfo{person}{Karen Simonyan}.}
  \bibinfo{year}{2020}\natexlab{}.
\newblock \showarticletitle{Off-policy actor-critic with shared experience
  replay}. In \bibinfo{booktitle}{\emph{International Conference on Machine
  Learning}}. PMLR, \bibinfo{pages}{8545--8554}.
\newblock


\bibitem[\protect\citeauthoryear{Schulman, Wolski, Dhariwal, Radford, and
  Klimov}{Schulman et~al\mbox{.}}{2017}]%
        {schulman2017proximal}
\bibfield{author}{\bibinfo{person}{John Schulman}, \bibinfo{person}{Filip
  Wolski}, \bibinfo{person}{Prafulla Dhariwal}, \bibinfo{person}{Alec Radford},
  {and} \bibinfo{person}{Oleg Klimov}.} \bibinfo{year}{2017}\natexlab{}.
\newblock \showarticletitle{Proximal policy optimization algorithms}.
\newblock \bibinfo{journal}{\emph{arXiv preprint arXiv:1707.06347}}
  (\bibinfo{year}{2017}).
\newblock


\bibitem[\protect\citeauthoryear{Stork, Zaefferer, and Bartz-Beielstein}{Stork
  et~al\mbox{.}}{2019a}]%
        {Stor18c}
\bibfield{author}{\bibinfo{person}{J{\"o}rg Stork}, \bibinfo{person}{Martin
  Zaefferer}, {and} \bibinfo{person}{Thomas Bartz-Beielstein}.}
  \bibinfo{year}{2019}\natexlab{a}.
\newblock \showarticletitle{Improving NeuroEvolution Efficiency by Surrogate
  Model-Based Optimization with Phenotypic Distance Kernels}. In
  \bibinfo{booktitle}{\emph{Applications of Evolutionary Computation}},
  \bibfield{editor}{\bibinfo{person}{Paul Kaufmann} {and}
  \bibinfo{person}{Pedro~A. Castillo}} (Eds.). \bibinfo{publisher}{Springer
  International Publishing}, \bibinfo{address}{Cham},
  \bibinfo{pages}{504--519}.
\newblock
\showISBNx{978-3-030-16692-2}


\bibitem[\protect\citeauthoryear{Stork, Zaefferer, Bartz-Beielstein, and
  Eiben}{Stork et~al\mbox{.}}{2019b}]%
        {Stor18d}
\bibfield{author}{\bibinfo{person}{J{\"o}rg Stork}, \bibinfo{person}{Martin
  Zaefferer}, \bibinfo{person}{Thomas Bartz-Beielstein}, {and}
  \bibinfo{person}{AE Eiben}.} \bibinfo{year}{2019}\natexlab{b}.
\newblock \showarticletitle{Surrogate models for enhancing the efficiency of
  neuroevolution in reinforcement learning}. In
  \bibinfo{booktitle}{\emph{Proceedings of the genetic and evolutionary
  computation conference}}. \bibinfo{pages}{934--942}.
\newblock


\bibitem[\protect\citeauthoryear{Stork, Zaefferer, Bartz-Beielstein, and
  Eiben}{Stork et~al\mbox{.}}{2020}]%
        {Stor20a}
\bibfield{author}{\bibinfo{person}{J{\"o}rg Stork}, \bibinfo{person}{Martin
  Zaefferer}, \bibinfo{person}{Thomas Bartz-Beielstein}, {and}
  \bibinfo{person}{AE Eiben}.} \bibinfo{year}{2020}\natexlab{}.
\newblock \showarticletitle{Understanding the Behavior of Reinforcement
  Learning Agents}. In \bibinfo{booktitle}{\emph{International Conference on
  Bioinspired Methods and Their Applications}}. \bibinfo{publisher}{Springer},
  \bibinfo{pages}{148--160}.
\newblock


\bibitem[\protect\citeauthoryear{Sutton and Barto}{Sutton and Barto}{2018}]%
        {sutton2018reinforcement}
\bibfield{author}{\bibinfo{person}{Richard~S Sutton} {and}
  \bibinfo{person}{Andrew~G Barto}.} \bibinfo{year}{2018}\natexlab{}.
\newblock \bibinfo{booktitle}{\emph{Reinforcement learning: An introduction}}.
\newblock \bibinfo{publisher}{MIT press}.
\newblock


\bibitem[\protect\citeauthoryear{Turner and Miller}{Turner and Miller}{2015}]%
        {turner2015introducing}
\bibfield{author}{\bibinfo{person}{Andrew~James Turner} {and}
  \bibinfo{person}{Julian~Francis Miller}.} \bibinfo{year}{2015}\natexlab{}.
\newblock \showarticletitle{Introducing a cross platform open source cartesian
  genetic programming library}.
\newblock \bibinfo{journal}{\emph{Genetic Programming and Evolvable Machines}}
  \bibinfo{volume}{16}, \bibinfo{number}{1} (\bibinfo{year}{2015}),
  \bibinfo{pages}{83--91}.
\newblock


\bibitem[\protect\citeauthoryear{Van~Beers and Kleijnen}{Van~Beers and
  Kleijnen}{2003}]%
        {van2003kriging}
\bibfield{author}{\bibinfo{person}{Wim~CM Van~Beers} {and}
  \bibinfo{person}{Jack~PC Kleijnen}.} \bibinfo{year}{2003}\natexlab{}.
\newblock \showarticletitle{Kriging for interpolation in random simulation}.
\newblock \bibinfo{journal}{\emph{Journal of the Operational Research Society}}
  \bibinfo{volume}{54}, \bibinfo{number}{3} (\bibinfo{year}{2003}),
  \bibinfo{pages}{255--262}.
\newblock


\bibitem[\protect\citeauthoryear{Wessing and Preuss}{Wessing and
  Preuss}{2017}]%
        {wessing2017true}
\bibfield{author}{\bibinfo{person}{Simon Wessing} {and} \bibinfo{person}{Mike
  Preuss}.} \bibinfo{year}{2017}\natexlab{}.
\newblock \showarticletitle{The true destination of EGO is multi-local
  optimization}. In \bibinfo{booktitle}{\emph{2017 IEEE Latin American
  Conference on Computational Intelligence (LA-CCI)}}. IEEE,
  \bibinfo{pages}{1--6}.
\newblock


\bibitem[\protect\citeauthoryear{Zaefferer}{Zaefferer}{2017}]%
        {CEGOv2.2.0}
\bibfield{author}{\bibinfo{person}{Martin Zaefferer}.}
  \bibinfo{year}{2017}\natexlab{}.
\newblock \bibinfo{title}{Combinatorial Efficient Global Optimization in {R} -
  {CEGO} v2.2.0}.
\newblock \bibinfo{howpublished}{online:
  https://cran.r-project.org/package=CEGO}.
\newblock
\newblock
\shownote{accessed: 2018-01-10.}


\bibitem[\protect\citeauthoryear{Zaefferer, Stork, Friese, Fischbach, Naujoks,
  and Bartz-Beielstein}{Zaefferer et~al\mbox{.}}{2014}]%
        {Zaefferer2014b}
\bibfield{author}{\bibinfo{person}{Martin Zaefferer}, \bibinfo{person}{J\"{o}rg
  Stork}, \bibinfo{person}{Martina Friese}, \bibinfo{person}{Andreas
  Fischbach}, \bibinfo{person}{Boris Naujoks}, {and} \bibinfo{person}{Thomas
  Bartz-Beielstein}.} \bibinfo{year}{2014}\natexlab{}.
\newblock \showarticletitle{Efficient Global Optimization for Combinatorial
  Problems}. In \bibinfo{booktitle}{\emph{Proc. GECCO'14}} (Vancouver, BC,
  Canada). \bibinfo{publisher}{ACM}, \bibinfo{pages}{871--878}.
\newblock


\end{thebibliography}

\end{document}